\documentclass[letterpaper, 10 pt, conference]{ieeeconf}

\IEEEoverridecommandlockouts
\usepackage{xcolor}         
\usepackage{etoolbox}       

\let\oldcite\cite
\renewcommand{\cite}[2][]{%
  \ifstrempty{#2}{%
    []%
  }{%
    \ifstrequal{#2}{?}{%
      []%
    }{%
      \ifstrempty{#1}{\oldcite{#2}}{\oldcite[#1]{#2}}%
    }%
  }%
}
\newcommand{\myparagraph}[1]{\noindent\textbf{#1}}

\usepackage[export]{adjustbox}
\usepackage{graphicx}   
\usepackage{subcaption}

\usepackage{float} 
\usepackage{placeins}

\usepackage{algorithm}
\usepackage{algorithmic}

\usepackage{booktabs}
\usepackage{utfsym}

\usepackage{wrapfig}
\usepackage{multirow}
\usepackage{multicol}
\usepackage{amsmath}
\usepackage{nicefrac}
\usepackage{amssymb}
\usepackage{bm}
\usepackage{bbm}

\renewcommand{\paragraph}[1]{\noindent\textbf{#1}}
\newcommand{\fref}[1]{Fig.~\ref{#1}}
\newcommand{\sref}[1]{Section~\ref{#1}}
\newcommand{\tref}[1]{Table~\ref{#1}}

\usepackage{balance}

\title{\LARGE \bf
Point-Level Refinable Semantic Mapping for Spatial Understanding
}

\author{Taimeng Fu$^{1*}$, Yi Du$^{1*}$, Zitong Zhan$^{1}$, Qiwei Du$^{1}$, Shaoshu Su$^{1}$, Zhipeng Zhao$^{1}$, and Chen Wang$^{1}$%
\thanks{$^{1}$Spatial AI \& Robotics Lab, Department of Computer Science and Engineering, University at Buffalo, USA.}
\thanks{$^{*}$Equal contribution.}%
}

\begin{document}

\maketitle
\thispagestyle{empty}
\pagestyle{empty}

\begin{abstract}
Open-vocabulary 3D semantic maps have emerged as a key spatial memory representation for long-horizon robot autonomy, enabling robots to reason about objects, their spatial relationships, and past observations across time. However, existing object-centric semantic maps are primarily designed for static environments and rely on an append-only point integration strategy, where newly observed points are accumulated but previously integrated points are rarely reevaluated. Consequently, perception errors accumulate over time, and outdated points persist after objects move or disappear. Although recent dynamic mapping methods update object-level representations in changing scenes, they do not explicitly revise noisy or obsolete points within each object. We propose an online refinable semantic mapping framework that continuously refines object representations through probabilistic point-level confidence updating. Each object point maintains a confidence score that is updated online using multi-view temporal evidence: consistent observations reinforce reliable points, while inconsistent segmentation masks, depth measurements, and visibility cues suppress spurious and outdated points. Combined with object tracking and re-identification to preserve instance identities under motion and relocation, our framework enables object-centric semantic maps to be continuously refined rather than monotonically accumulated. Experiments demonstrate improved mapping accuracy over the previous method by 3.46\% on Replica and 2.55\% on ScanNet, show robustness in dynamic environments and under perception noise, and provide an effective spatial memory for object navigation and embodied question-answering with success rates of 82.4\% and 76.7\%.

\end{abstract}

\section{Introduction}
\label{sec:intro}

Open-vocabulary 3D semantic maps have emerged as a key spatial memory representation for robot autonomy \cite{wang2025imperative} by grounding open-ended semantic concepts in 3D space \cite{rotondi20263d}. They enable robots to localize task-relevant objects, reason about object relationships, and maintain persistent spatial context for embodied tasks, e.g., vision-language navigation~\cite{zhang2024vision} and mobile manipulation~\cite{du2026iflax}.

Recent open-vocabulary mapping systems~\cite{gu2024conceptgraphs, Werby-RSS-24, maggio2024clio} construct object-centric 3D semantic maps by integrating RGB-D observations with open-vocabulary perception models. However, real-world robotic deployment requires long-term operation in dynamic environments, where objects are frequently moved, removed, or rearranged. Existing methods are largely designed for static scenes and rely on an append-only point integration strategy: newly observed points are continuously added to object instances, while previously integrated points are rarely revisited or revised.

\begin{figure}[t]
    \centering
    \includegraphics[width=0.9\linewidth]{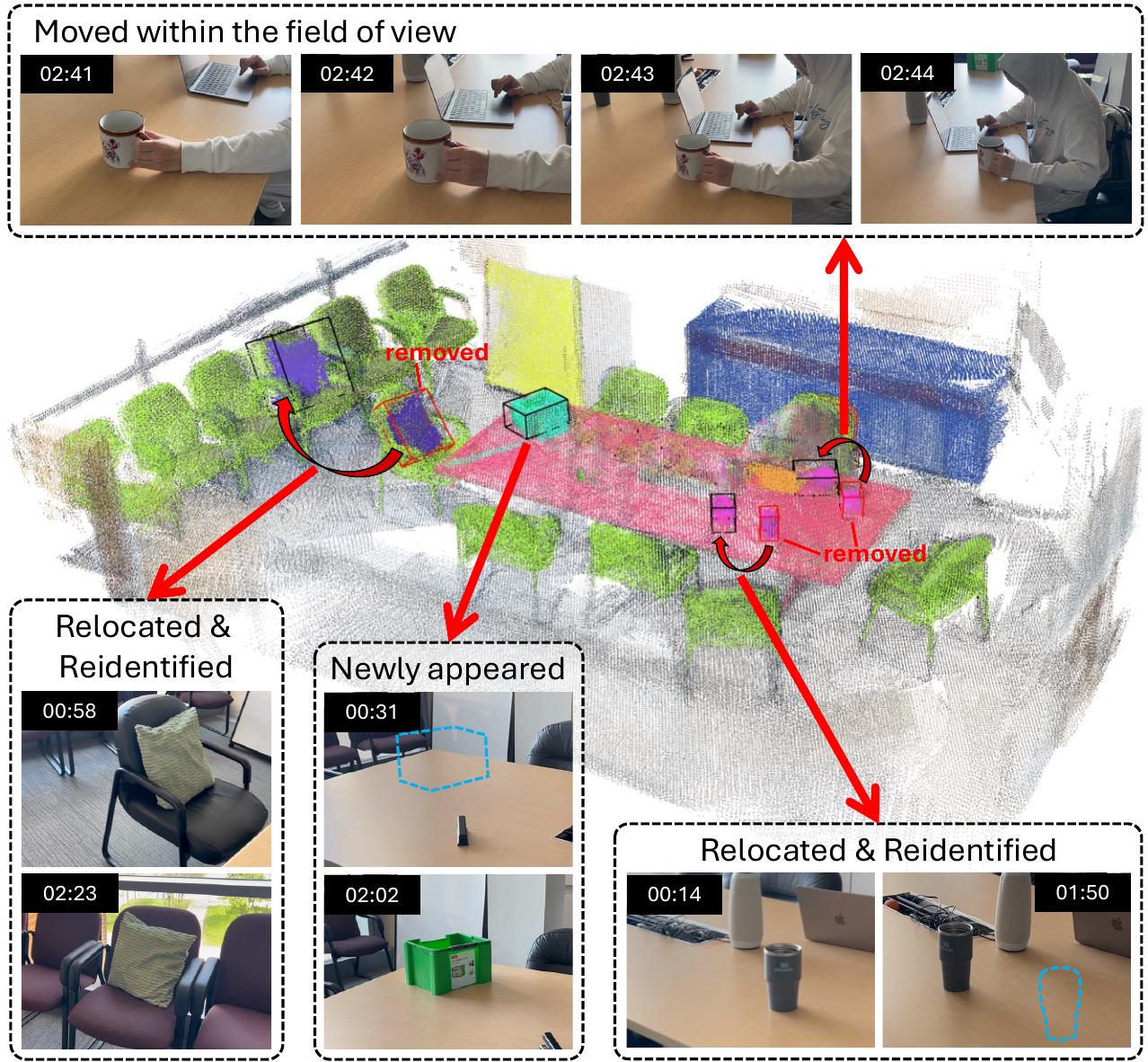} 
    \caption{
    \textbf{Dynamic open-vocabulary semantic mapping.} Our method maintains a refinable object-centric 3D semantic map through point-level probabilistic updates. It handles dynamic objects that newly appear, move within the robot's view, or are relocated out of view and later re-identified.
    }
    \label{fig:head}
\end{figure}

Several recent methods have extended semantic mapping to dynamic environments by detecting object changes and updating object-level map representations~\cite{Schmid-RSS24-Khronos,tang2025openin,jiang2025dualmap}. For example, Khronos~\cite{Schmid-RSS24-Khronos} identifies changed objects through depth consistency, while OpenIN~\cite{tang2025openin} and DualMap~\cite{jiang2025dualmap} associate movable objects with supporting anchors to maintain object identities. However, these approaches primarily reason about scene changes at the object or relation level, without explicitly revising noisy or outdated points within each object instance. As a result, local artifacts introduced by perception errors or stale observations can remain embedded in the object representation, degrading long-term mapping quality.

To address both dynamic environments and noisy perception, we propose an online refinable open-vocabulary semantic mapping framework that continuously refines object-centric maps over time. The key idea is to maintain a probabilistic confidence estimate for every object point, rather than treating integrated points as permanent entries. Each confidence score is updated online using multi-view temporal evidence: points consistently supported by new observations are reinforced, while those inconsistent with segmentation masks, depth measurements, or visibility cues are progressively down-weighted and eventually removed. This enables semantic maps to preserve reliable geometry while eliminating perception artifacts and outdated points caused by object motion. To maintain consistent object identities, the framework further combines 2D tracking for in-view motion with feature-based re-identification after objects leave and re-enter the field of view, as shown in \fref{fig:head}.

To evaluate its effectiveness as an online spatial memory for autonomous robots, we integrate the proposed semantic mapping framework into a vision-language navigation (VLN) framework \cite{du2026vlnav} and evaluate it on object navigation and embodied question answering tasks in real-world environments.
The main contributions of this work are:
\begin{itemize}
\item We propose an online refinable semantic mapping framework with point-level confidence updating, enabling object-centric semantic maps to be continuously refined rather than monotonically accumulated.
\item We introduce a multi-view temporal update mechanism that reinforces reliable object points while suppressing spurious and outdated points caused by perception errors and object motion, improving mapping accuracy and robustness in dynamic environments.
\item We integrate object tracking and feature-based re-identification to preserve consistent object identities under motion and demonstrate the proposed map as an effective online spatial memory for vision-language navigation and embodied question answering.
\end{itemize}

\section{Related Work}
\label{sec:related}

\subsection{Open-Vocabulary Semantic Mapping}

Semantic mapping augments geometric reconstruction with object labels, instance structure, and region-level semantics for navigation and manipulation \cite{rotondi20263d}. Earlier metric-semantic systems \cite{hughes2022hydra} rely on closed-set detectors \cite{redmon2016you}, while recent vision-language foundation models \cite{Cheng2024YOLOWorld, carion2025sam} enable open-vocabulary maps that can be queried with free-form language.
Building on these advances, recent works integrate open-vocabulary semantic maps with large language models (LLMs) for embodied reasoning and long-horizon planning \cite{gu2024conceptgraphs, Werby-RSS-24, maggio2024clio}.
These approaches demonstrate the value of structured semantic memory for language-guided robotic reasoning. However, they mainly focus on semantic abstraction and typically assume that fused object points remain valid once inserted. As a result, erroneous or old points introduced by noisy perception or object motion are difficult to revise through later observations.

\subsection{Dynamic Semantic Map Updating}

Real robotic environments are dynamic, where task-relevant objects may be moved, removed, or re-observed at different locations.
Dynamic semantic mapping systems reason about environmental changes using geometric consistency and object relations.
Khronos \cite{Schmid-RSS24-Khronos} uses ray-casting to verify geometric changes, but its verification is primarily object-level and does not explicitly refine erroneous points within an otherwise valid instance or suppress false semantic instances using semantic inconsistency. OpenIN \cite{tang2025openin} and DualMap \cite{jiang2025dualmap} maintain movable objects through relations to relatively stable anchors. Such relation-based updating depends on the existence of a suitable anchor and sufficiently complete re-observation of that anchor. In contrast, our method maintains each instance independently without requiring predefined object-to-anchor relations.

\begin{figure*}[t]
    \centering
    \includegraphics[width=0.9\linewidth]{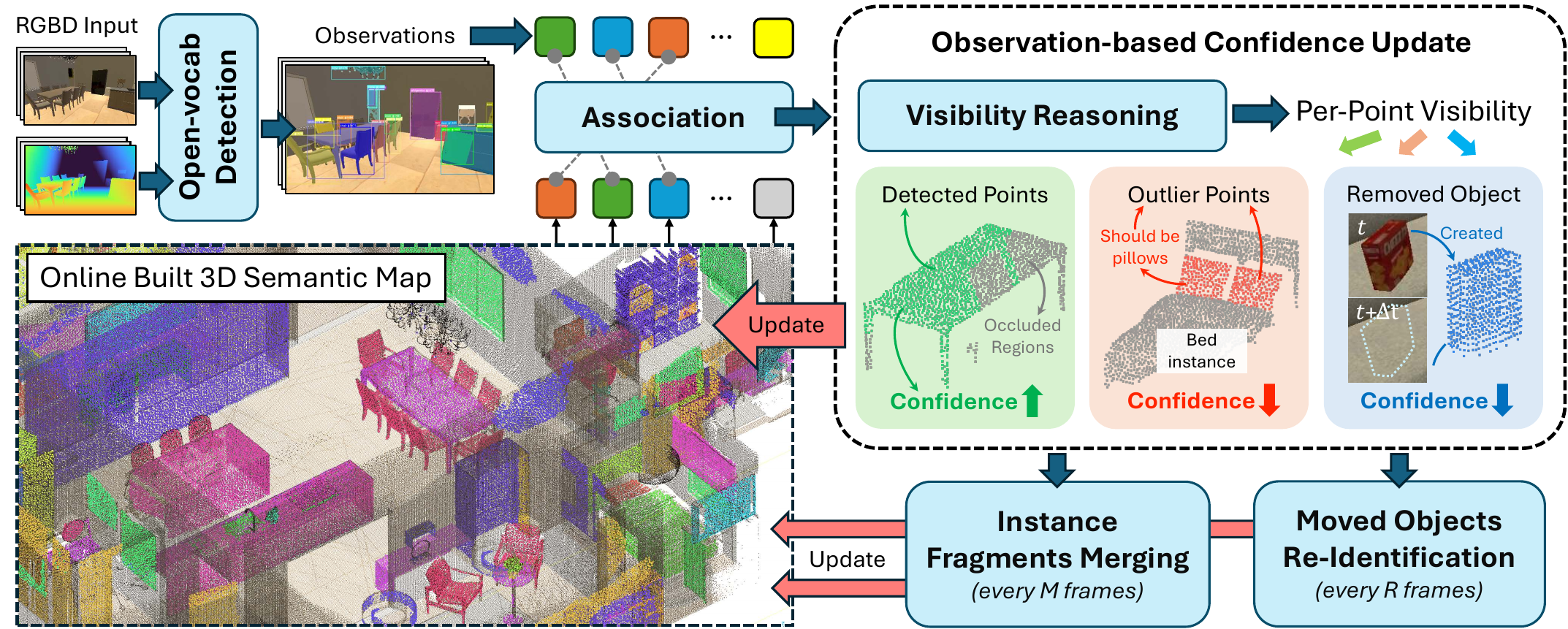} 
    \caption{
    \textbf{Overview of the proposed dynamic semantic mapping framework.}
    Given posed RGB-D observations, open-vocabulary detections are generated and associated with object instances in the 3D semantic map.
    For each association, we perform a visibility-based point confidence update.
    Points supported by current detections, such as the green points on the table, receive higher confidence.
    In contrast, inconsistent outlier points, such as red points mistakenly fused into a bed instance due to previous segmentation errors, are down-weighted when later observations identify them as pillows.
    Points belonging to removed objects are identified through visibility reasoning, assigned lower confidence, and eventually pruned.
    Unobserved points in the current frame, shown in gray, remain unchanged until more evidence is collected in later frames.
    The semantic map is updated online in this frame-by-frame manner, while periodic instance-fragment merging reduces duplicate instances and re-identification preserves object identities when objects are relocated and re-observed at new locations.
    }
    \label{fig:flow}
\end{figure*}

\section{Method}
\label{sec:method}

Given RGB-D streams, our goal is to build an online open-vocabulary 3D semantic map that uses each observation to both add new object points and revise existing ones through detection and visibility cues, as illustrated in \fref{fig:flow}.

\subsection{Probabilistic Map Representation}
\label{sec:map_repr}

We represent the 3D semantic map as a set of object instances, where each instance contains 3D points with probabilistic confidence scores for online point-level revision.
\begin{equation}
\mathcal{M} = \{o_i: (\text{id}_i,\mathcal{P}_i,\mathcal{S}_i,\mathbf{h}_i,\text{bb}_i,\mathbf{f}_i^{\text{clip}},\mathbf{f}_i^{\text{dino}},n_i^{\text{obs}},t_i^{\text{obs}}) \},
\end{equation}
where $\mathcal{P}_i$ and $\mathcal{S}_i$ denote object points and their confidence scores; $\mathbf{h}_i$, $\text{bb}_i$, $n_i^{\text{obs}}$, and $t_i^{\text{obs}}$ store label histogram, bounding box, number and the last timestamp of observations; $\mathbf{f}_i^{\text{clip}}$, $\mathbf{f}_i^{\text{dino}}$ are descriptors used for retrieval and re-identification.

The per-point score $s_{ij}$ represents the confidence of point $p_{ij}$ and is updated probabilistically from observations.
To achieve efficient probabilistic update, inspired by \cite{hornung13auro}, we define score as log-odds of its confidence probability $c_{ij}$:
\begin{equation}
s_{ij} = \sigma^{-1}(c_{ij}) = \log \frac{c_{ij}}{1-c_{ij}}.
\label{eq:log_odds}
\end{equation}
This log-odds formulation allows the score increment from new observations to be directly added to the existing score, which is equivalent to Bayesian probability updating but more computationally efficient.

\subsection{Online Semantic Mapping}
\label{sec:update}

\subsubsection{Open-vocabulary Detection}
Denote the RGB image, depth image, and camera pose at time $t$ as $(\mathbf{I}_t,\mathbf{D}_t,\mathbf{T}_t)$. We employ vision foundation models to predict open-vocabulary detections and segmentation masks:
\begin{equation}
\{ \mathbf{m}_k, l_k, c_k, \mathbf{g}_k^{\text{clip}}, \mathbf{g}_k^{\text{dino}} \}_{k=1}^{n_t} = \text{Perception}(\mathbf{I}_t, \text{prompt}),
\end{equation}
where $\mathbf{m}_k$, $l_k$, and $c_k$ are the 2D mask, semantic label, and detection confidence of the $k$-th detection, respectively. $\mathbf{g}_k^{\text{clip}}$ and $\mathbf{g}_k^{\text{dino}}$ denote the CLIP \cite{vasu2024mobileclip} and DINO \cite{simeoni2025dinov3} features extracted from the detected region. The $\text{prompt}$ is an open-vocabulary class list generated by an LLM.

With pixel-aligned depth $\mathbf{D}_t$ and camera pose $\mathbf{T}_t$, each 2D detection is lifted to a 3D point cloud in the world frame:
\begin{equation}
\mathcal{P}_k = \Pi^{-1}(\mathbf{m}_k,\mathbf{D}_t;\mathbf{K},\mathbf{T}_t),
\end{equation}
where $\Pi^{-1}$ denotes masked RGB-D back-projection and $\mathbf{K}$ is the camera intrinsic matrix. 
To filter out depth noise and reconstruction artifacts, we apply DBSCAN \cite{ester1996density} to $\mathcal{P}_k$ and only retain the largest connected component.

\subsubsection{Point-level Visibility Reasoning}
\label{sec:vis_update}
An object that is not detected in the current frame should not be immediately treated as nonexistent, since it may lie outside the field of view, be occluded by other objects, or be only partially observed. We therefore perform point-level visibility reasoning to determine how each map point relates to the current observation, which provides the basis for subsequent confidence updates.
To this end, we project each object point onto the current camera view:
\begin{equation}
(\mathbf{u}_{ij}, d_{ij})=\Pi(\mathbf{p}_{ij};\mathbf{K},\mathbf{T}_t),
\end{equation}
where $\mathbf{u}_{ij}$ is the projected pixel coordinate and $d_{ij}$ is the point depth. We only consider points whose projections fall within the image bounds and satisfy $d_{ij}>0$. By comparing $d_{ij}$ with the observed pixel depth $\mathbf{D}_t(\mathbf{u}_{ij})$, each point is assigned a visibility label following:
\begin{equation}
v_{ij} =
\begin{cases}
\mathrm{visible}, & |d_{ij}-D_t(\mathbf{u}_{ij})|\leq\epsilon_d,\\
\mathrm{missed}, & d_{ij}<D_t(\mathbf{u}_{ij})-\epsilon_d,\\
\mathrm{occluded}, & \mathrm{otherwise},
\end{cases}
\label{eq:visibility}
\end{equation}
where $\epsilon_d$ is a small depth tolerance.
The label $\mathrm{visible}$ indicates that the point lies on the observed surface and can be revised using detection evidence. The label $\mathrm{missed}$ indicates that the point is unsupported by the current depth measurement, suggesting it may be removed. The label $\mathrm{occluded}$ indicates that the point is behind the observed surface and should remain unchanged in the current frame.

In practice, we perform visibility reasoning and subsequent score updating only on a local subset of instances in the map within a predefined distance $d_{\mathrm{th}}$ from the camera. This restricts computation to nearby objects and prevents the cost from growing with the global map.
For each local instance, we denote its visible point subset and corresponding scores as $\mathcal{P}_i^{\mathrm{visible}}$ and $\mathcal{S}_i^{\mathrm{visible}}$, respectively, and use analogous notation for the $\mathrm{missed}$ and $\mathrm{occluded}$ subsets.

\subsubsection{Detection-to-Instance Association} \label{sec:asso}
We associate detections with existing instances, considering both semantic similarity and geometric overlap. For each detection-instance pair, we compute the semantic cosine similarity
$\phi_{ki} = \langle \mathbf{g}^{\mathrm{clip}}_k, \mathbf{f}^{\mathrm{clip}}_i \rangle$
and the 3D intersection-over-union (IoU) $\psi_{ki}$ between the detection's point cloud $\mathcal{P}_k$ and the visible object points $\mathcal{P}_i^{\mathrm{visible}}$. The matching score is
$\mathrm{Sim}_{ki}=\phi_{ki}+\psi_{ki}$, and the best match is accepted as
$i^*=\arg\max_i \mathrm{Sim}_{ki}$ if the maximum score exceeds a predefined threshold $\mathrm{Sim}_{\mathrm{th}}$. 

If a valid match is found, the detection is integrated into the matched instance by merging points and updating the label histogram and CLIP/DINO descriptors. If no match is found, a new instance is initialized.
After all detections in a frame have been processed, each updated object point cloud is voxel-downsampled to reduce redundancy, and its bounding box is recomputed from the resulting point cloud.

\subsubsection{Confidence Score Updating} \label{sec:score}

To enable point-level revision in dynamic scenes, we update the confidence scores of object points according to their visibility status and consistency with current observations.
The key principle is to treat each frame as evidence for both map reinforcement and refinement: points supported by current detections receive positive evidence, points contradicted by masks or depth measurements receive negative evidence, and unobserved points remain unchanged.
We consider four update cases:

\myparagraph{In-Mask Points}.
When an object instance $o_i$ is associated with a detection $k$, its inlier points are defined as the visible points whose reprojections fall inside the segmentation mask:
\begin{equation}
\mathcal{P}_i^{\text{inlier}}=\{\mathbf{p}_j : \mathbf{p}_j \in \mathcal{P}_i^{\text{visible}}, \Pi(\mathbf{p}_j;\mathbf{K}, \mathbf{T}_t) \in \mathbf{m}_k\}_{j=1}^{|\mathcal{P}_i^{\text{visible}}|}.
\end{equation}
These inlier points receive positive evidence from the current observation, so their confidence scores are increased by the log-odds of the corresponding detection confidence:
\begin{equation}
    \mathcal{S}_i^{\text{inlier}} \leftarrow \mathcal{S}_i^{\text{inlier}} + \sigma^{-1}(c_k).
\end{equation}

\myparagraph{Out-of-Mask Points}.
In contrast, the visible but out-of-mask points of the object (i.e., $\mathcal{P}_i^{\text{outlier}} = \mathcal{P}_i^{\text{visible}} - \mathcal{P}_i^{\text{inlier}}$) receive negative evidence from the current observation. We decrease their confidence scores by a predefined value:
\begin{equation}
    \mathcal{S}_i^{\text{outlier}} \leftarrow \mathcal{S}_i^{\text{outlier}} - \Delta_{\text{outlier}}.
\end{equation}

\myparagraph{Object Not Detected}. 
When an instance is not associated with any detection in the current frame, it may be either (1) a ``ghost'' instance introduced by previous erroneous detections, or (2) a valid object that is difficult to detect due to limited visibility, such as heavy occlusion or a small image footprint. To avoid penalizing occluded or weakly visible objects, we estimate detectability using a visibility ratio:
\begin{equation}
r=|\text{CloseOp}(\mathbf{u}_i^{\text{visible}})| / |\text{CloseOp}(\mathbf{u}_i)|,
\end{equation}
where $\mathbf{u}_i^{\text{visible}}$ and $\mathbf{u}_i$ are pixel coordinates of visible points and all points;
$\text{CloseOp}(\cdot)$ denotes a morphological closing operation that fills small holes in the reprojected region, and $|\cdot|$ denotes the number of pixels.
If the visible area ratio and image footprint exceed thresholds, i.e.,
\begin{equation}
(r \geq \tau_{\text{occ}}) \; \land \; (|\text{CloseOp}(\mathbf{u}_i^{\text{visible}})| \geq \tau_{\text{px}}),
\label{eq:should_det_check}
\end{equation}
the object should be detectable.
Absence of a matched detection indicates the instance is more likely to be a ``ghost'' introduced by previous erroneous detections, and we therefore reduce the scores of all points by $\Delta_{\neg\text{det}}$.

\myparagraph{Depth-Contradicted Points}.
Points classified as \textrm{missed} in \eqref{eq:visibility} contradict the current depth observation, as they lie in front of the observed surface along the same viewing rays. This indicates that these points are no longer supported at their mapped locations, which may occur when an object has been removed from its previous position or when depth errors have introduced spurious points. We therefore decrease these points' confidence scores by $\Delta_{\text{missed}}$.

After the confidence update for the current frame, points with scores below $s_{\min}$ are pruned, and instances with fewer than $n_{\min}$ points are removed.
This confidence updating and pruning procedure preserves stable, consistently supported object points while gradually eliminating old or spurious points caused by perception noise or object motion.
As a result, the map maintains persistent object instances while correcting their local geometry online in dynamic scenes.

\subsection{Object Fragment Merging}
\label{sec:optimization}

Object fragmentation occurs when disconnected partial observations of the same physical object are initialized as separate instances.
To reduce such fragmentation, we periodically merge instances once sufficient geometric and visual cues indicate that they correspond to the same object.
Specifically, for two objects $o_i,o_j$, we merge them if
\begin{equation}
\begin{aligned}
\langle \mathbf{f}_i^{\text{clip}}, \mathbf{f}_j^{\text{clip}} \rangle &\geq \tau_{\text{clip}}^{\text{frag}}, \\
\langle \mathbf{f}_i^{\text{dino}}, \mathbf{f}_j^{\text{dino}} \rangle &\geq \tau_{\text{dino}}^{\text{frag}}, \\
\mathrm{Overlap}(\mathcal{P}_i,\mathcal{P}_j) &\geq \tau_{\text{overlap}}^{\text{frag}},
\end{aligned}
\label{eq:fragment}
\end{equation}
where $\mathrm{Overlap}(\cdot)$ measures the 3D point cloud overlap ratio following \cite{gu2024conceptgraphs}, and $\tau_{\text{clip}}^{\text{frag}}$, $\tau_{\text{dino}}^{\text{frag}}$, and $\tau_{\text{overlap}}^{\text{frag}}$ are thresholds.
This conservative merging rule requires agreement in geometry, semantics, and appearance, thereby effectively correcting object fragmentation while avoiding false merges.

\subsection{Dynamic Object Tracking and Re-Identification}
\label{sec:reid}

Downstream tasks such as VLN and manipulation require stable instance identities even after objects move. We therefore use 2D tracking for in-view object motion and feature-based re-identification for out-of-view object relocation to maintain consistent instance IDs.

\subsubsection{Tracking Moving Objects in the Camera View}
For real-time efficiency, we use BoT-SORT \cite{aharon2022bot}, a lightweight 2D tracker, to associate detections across frames.
When a new detection is first observed, it is matched to an existing instance or a new one is created according to \sref{sec:asso}. Subsequent tracked detections with the same tracking ID are directly fused into this instance.
This prevents a moving object from being repeatedly initialized as new instances and preserves its instance ID during the motion.
Trailing points left at previous locations are removed by the confidence update mechanism (\sref{sec:score}) within the next 1-2 frames, as they contradict later depth observations.

\subsubsection{Re-Identification of Relocated Objects}
When an object relocates outside the camera's field of view, its motion cannot be directly tracked. The system must re-identify the object when it is rediscovered. To this end, we integrate a feature-based re-identification mechanism into semantic mapping.

Specifically, when an instance $o_i$ is removed from the map because of depth contradiction, we store a compact record containing its instance ID, bounding box, and visual features:
\begin{equation}
o_i^\prime=(\text{id}_i,\text{bb}_i,\mathbf{f}^{\text{clip}}_i,\mathbf{f}^{\text{dino}}_i).
\end{equation}
The record is inserted into buffer $\mathcal{B}$ with an importance score:
\begin{equation}
\mathcal{B}=\{(\lambda_i,o_i^\prime)\},\quad
\lambda_i=t_i^{\text{obs}} + 10\times n_i^{\text{obs}},
\end{equation}
where $\lambda_i$ is the importance score that prioritizes recently and frequently observed objects. When the buffer exceeds its capacity, entries with the lowest scores are discarded.
The re-identification is triggered periodically in a standalone process. A removed object $o_r^\prime \in \mathcal{B}$ and a current instance in the map $o_i \in \mathcal{M}$ are considered a match if
\begin{equation}
\begin{gathered}
\langle \mathbf{f}^{\text{clip}}_r,\mathbf{f}^{\text{clip}}_i\rangle \geq \tau^{\text{reid}}_{\text{clip}},\\
\langle \mathbf{f}^{\text{dino}}_r,\mathbf{f}^{\text{dino}}_i\rangle \geq \tau^{\text{reid}}_{\text{dino}},\\
|\text{Vol}(\text{bb}_r) - \text{Vol}(\text{bb}_i)| \leq \tau^{\text{reid}}_{\text{vol}},
\end{gathered}
\label{eq:reid}
\end{equation}
where $\text{Vol}(\cdot)$ denotes the 3D bounding box volume. The thresholds $\tau^{\text{reid}}_{\text{clip}}$, $\tau^{\text{reid}}_{\text{dino}}$, and $\tau^{\text{reid}}_{\text{vol}}$ enforce semantic, appearance, and geometric size consistency.
If the removed object $o_r^\prime$ matches multiple instances in the map, such as a removed chair matching several identical chairs in an office, we treat its identity as ambiguous and skip re-identification. In contrast, if $o_r^\prime$ is uniquely and consistently matched to the same object $o_i$ for three consecutive re-identification attempts, we regard it as a confident match and assign the stored ID $\mathrm{id}_r$ to $o_i$.
This process is independent of observation order: an object may first disappear and later be re-identified, or it may first appear as a new instance and be re-identified after its removal from the old location is confirmed.

\begin{table}[!t]\centering\small
\caption{Evaluation results of open-vocabulary 3D semantic mapping methods on Replica and ScanNet datasets.}\label{tab:mapping}
\begin{tabular}{llccc}\toprule
\textbf{Dataset} &\textbf{Method} &\textbf{mIoU} &\textbf{F-mIoU} &\textbf{mAcc} \\\midrule
\multirow{4}{*}{Replica} &ConceptGraphs &0.1501 &0.3858 &0.3559 \\
&HOV-SG &0.2050 &0.4846 &0.3835 \\
&DualMap &0.2538 &0.5207 &0.4024 \\
&\textbf{Ours} &\textbf{0.2629} &\textbf{0.5530} &\textbf{0.4671} \\\midrule
\multirow{4}{*}{ScanNet} &ConceptGraphs &0.0882 &0.3077 &0.3538 \\
&HOV-SG &0.1333 &0.3381 &0.3714 \\
&DualMap &0.1604 &0.3288 &0.3794 \\
&\textbf{Ours} &\textbf{0.1646} &\textbf{0.3405} &\textbf{0.3850} \\
\bottomrule
\end{tabular}
\end{table}

\begin{table}[!t]
\centering
\small
\caption{Computational efficiency and final map size.}
\label{tab:efficiency}
\resizebox{\linewidth}{!}{
\begin{tabular}{llcccccc}
\toprule
\multirow{2}{*}{\textbf{Dataset}} &
\multirow{2}{*}{\textbf{Method}} &
\multicolumn{4}{c}{\textbf{Memory (MB)}} &
\multirow{2}{*}{\shortstack{\textbf{TPF (s)}}} &
\multirow{2}{*}{\shortstack{\textbf{Avg. Points}\\\textbf{/ Scene}}} \\
\cmidrule(lr){3-6}
& & \textbf{CPU Avg.} & \textbf{CPU Peak} & \textbf{GPU Avg.} & \textbf{GPU Peak} & & \\
\midrule
\multirow{2}{*}{Replica}
& DualMap & 5391.5 & 7813.3 & 7921.0 & 12248.0 & 0.2219 & 255,234 \\
& \textbf{Ours} & \textbf{4492.6} & \textbf{4966.7} & \textbf{5136.4} & \textbf{7002.0} & \textbf{0.1811} & 209,617 \\
\midrule
\multirow{2}{*}{ScanNet}
& DualMap & 4668.5 & 6027.9 & 6596.8 & 8608.0 & 0.1544 & 218,785 \\
& \textbf{Ours} & \textbf{4651.4} & \textbf{5835.7} & \textbf{5766.6} & \textbf{7574.0} & \textbf{0.1324} & 194,454 \\
\bottomrule
\end{tabular}
}
\end{table}

\begin{table}[!t]\centering\small
\caption{Ablation of point-level confidence updating and fragment merging on the Replica dataset.}\label{tab:abl_score_merge}
\resizebox{\linewidth}{!}{
\begin{tabular}{lcccc}\toprule
\textbf{Method} &\textbf{mIoU} &\textbf{F-mIoU} &\textbf{mAcc} &\textbf{Duplication} \\\midrule
w/o confidence update &0.2529 &0.5384 &0.4629 &1.2568 \\
w/o fragment merging &0.2502 &0.5479 &0.4539 &1.3426 \\
\textbf{Ours} &\textbf{0.2629} &\textbf{0.5530} &\textbf{0.4671} &\textbf{1.1690} \\
\bottomrule
\end{tabular}}
\end{table}

\subsection{Integration with VLN framework}
\label{sec:vln}

We integrate the proposed semantic mapping method into VL-Nav \cite{du2026vlnav} as a compact symbolic-spatial memory for LLM-based robotic agents.
In this framework, Qwen2.5-VL 7B \cite{bai2025qwen25vl} serves as the central coordinator, interpreting language instructions, querying the semantic map for object layouts, and invoking skills such as object navigation and embodied question answering (EQA).
The semantic map is constructed online during robot operation to maintain up-to-date scene information. It supports fuzzy semantic queries by computing CLIP similarity between query terms generated by the LLM and stored instance features. The query returns metadata of the top-$K$ matched objects in JSON format, which the LLM uses to plan skill execution. For object navigation, if the target exists in the map, the navigation skill invokes a planner to generate a path and navigate to it; otherwise, the robot performs frontier-based exploration to collect more observations. For EQA, the LLM reasons over the exploration progress and conversation history to decide whether sufficient information has been gathered to answer the question or whether further exploration is needed.

\section{Experiments}
\label{sec:experiments}

We evaluate mapping accuracy, component contributions, robustness to perception errors, dynamic-object handling, and downstream benefits for real-world VLN.

\subsection{Experimental Setup}

In practice, we adopt a perception pipeline similar to \cite{jiang2025dualmap}, which provides a good balance between accuracy and efficiency. Specifically, we use YOLO-World \cite{Cheng2024YOLOWorld} for open-vocabulary detection, MobileSAM \cite{mobile_sam} for object mask generation, and FastSAM \cite{zhao2023fast} to segment additional objects outside the prompt list. CLIP features are computed as a mixture of 70\% image and 30\% text embeddings using MobileCLIP \cite{vasu2024mobileclip}. DINO features are obtained by generalized mean pooling over patch features from the lightweight ViT-S+/16 model \cite{simeoni2025dinov3}.
Parameters for perception and association follow DualMap \cite{jiang2025dualmap}. Refinement parameters were calibrated on two held-out HSSD scenes, and re-identification thresholds on a set of independently collected object images. All values were then fixed across experiments. We use $\mathrm{Sim}_{\mathrm{th}}=1.2$, $\epsilon_d=3\,\mathrm{cm}$, $\Delta_{\text{outlier}}=0.2$, $\Delta_{\neg\text{det}}=0.12$, $\Delta_{\text{missed}}=0.85$, $\tau_{\mathrm{occ}}=0.5$, $\tau_{\mathrm{px}}=3000$, $(\tau_{\text{clip}}^{\text{frag}},\tau_{\text{dino}}^{\text{frag}},\tau_{\text{overlap}}^{\text{frag}})=(0.9,0.9,0.7)$, and $(\tau^{\text{reid}}_{\text{clip}},\tau^{\text{reid}}_{\text{dino}},\tau^{\text{reid}}_{\text{vol}})=(0.9,0.95,0.1)$.

\begin{table}[!t]\centering\small
\caption{Impact of simulated segmentation errors on semantic mapping performance on the Replica dataset.}\label{tab:drop}
\begin{tabular}{lcccc}\toprule
\textbf{Method} &$\boldsymbol\Delta$\textbf{mIoU} &$\boldsymbol\Delta$\textbf{F-mIoU} &$\boldsymbol\Delta$\textbf{mAcc} \\\midrule
DualMap &-11.62\% &-3.06\% &-10.13\% \\
\textbf{Ours} &\textbf{-6.29\%} &\textbf{-3.04\%} &\textbf{-7.77\%} \\
\bottomrule
\end{tabular}
\end{table}

\begin{figure}[!t]
    \centering
    \includegraphics[width=0.8\linewidth]{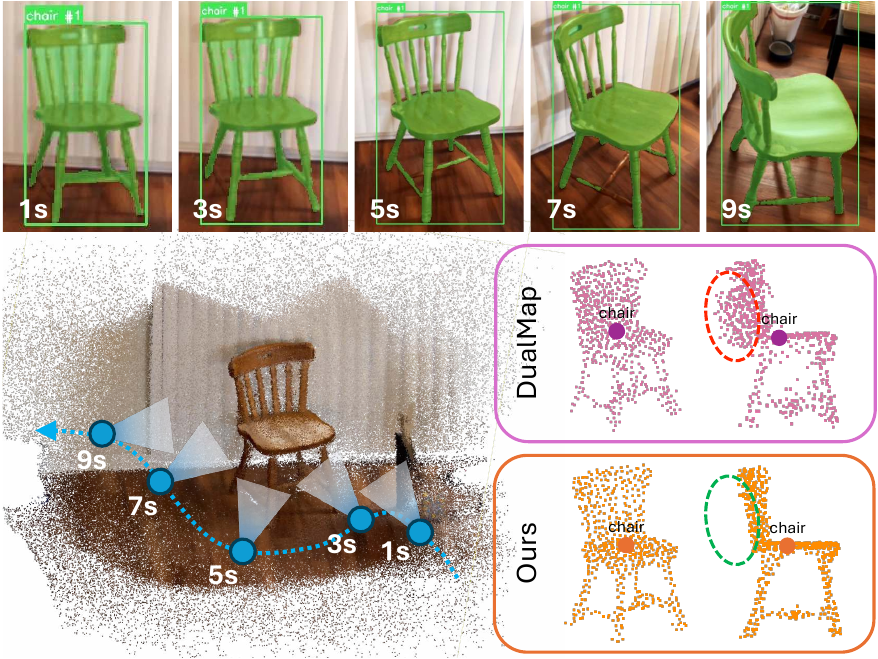}
    \caption{
    \textbf{Example of handling inaccurate masks.} A wooden chair with thin and hollow structures is observed from multiple viewpoints. Inaccurate masks may fill the gaps between chair parts, introducing outlier points when fused into 3D. DualMap retains these mask-induced artifacts, while our point-level confidence update suppresses inconsistent points across views and reconstructs cleaner chair geometry.
    }
    \label{fig:chair}
\end{figure}

\subsection{Mapping Accuracy}

To evaluate open-vocabulary 3D semantic mapping accuracy, we use the Replica \cite{replica19arxiv} and ScanNet \cite{dai2017scannet} benchmarks following the scene selection of \cite{jiang2025dualmap}. All methods are evaluated on the final reconstructed semantic map, excluding background-dominant classes such as wall, floor, and ceiling. We compare with three state-of-the-art open-vocabulary semantic mapping baselines: ConceptGraphs \cite{gu2024conceptgraphs}, HOV-SG \cite{Werby-RSS-24}, and DualMap \cite{jiang2025dualmap}. ConceptGraphs and HOV-SG are offline methods, while DualMap is online, like ours. We adopt the standard metrics from \cite{jiang2025dualmap}, including mean IoU (mIoU), frequency-weighted IoU (F-mIoU), and mean accuracy (mAcc).
The results are summarized in \tref{tab:mapping}. Our method outperforms all baselines on both datasets. Compared with DualMap, the previous state-of-the-art, our method improves mIoU by 3.46\% and 2.55\%, F-mIoU by 5.84\% and 3.44\%, and mAcc by 13.85\% and 1.45\% on Replica and ScanNet, respectively.

\begin{figure}[!t]
    \centering
    \includegraphics[width=0.9\linewidth]{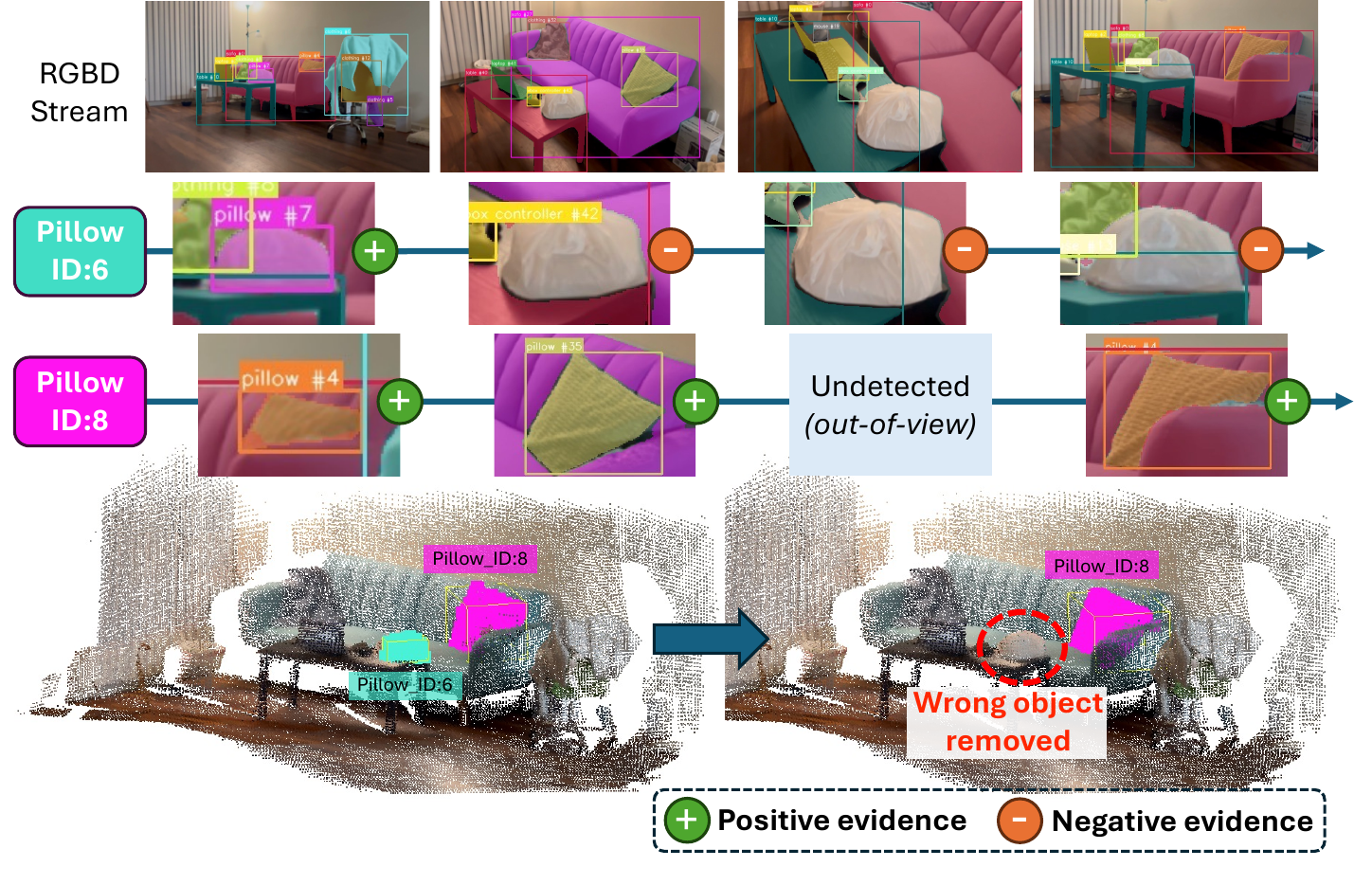}
    \caption{
    \textbf{Example of handling incorrect detections.} A plastic bag is mistakenly detected as a pillow, creating a false instance in the map. Our confidence update mechanism accumulates positive evidence for consistently detected objects and negative evidence for unsupported instances, eventually removing the false pillow while preserving the true one.
    }
    \label{fig:plastic}
\end{figure}

\begin{figure}[!t]
    \centering
    \includegraphics[width=0.85\linewidth]{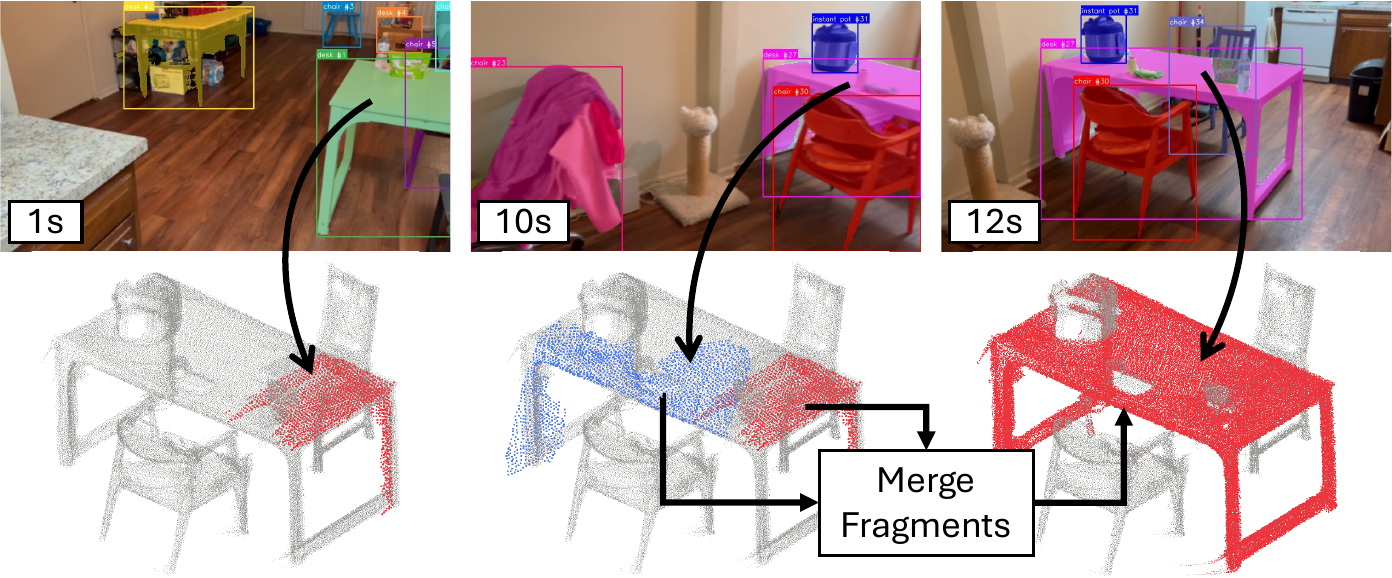}
    \caption{
    \textbf{Example of object fragment merging.} A table is initially reconstructed as two separate instances due to temporally disconnected partial observations at 1s and 10s. As more observations become available at 12s, our method merges the fragments into a complete object instance.
    }
    \label{fig:table}
\end{figure}

\subsection{Computational Efficiency}
Table~\ref{tab:efficiency} compares memory usage and runtime on the same hardware platform. Our method uses less CPU memory because incremental refinement does not require storing previous observation buffers, and less GPU memory by back-projecting only valid mask regions rather than expanding full-resolution depth tensors. Despite additional visibility reasoning and confidence updates, it achieves lower average time-per-frame (TPF) through implementation optimizations such as batched object processing, parallel feature extraction, and pipelined execution of perception and mapping.

\subsection{Controlled Ablations}

We isolate the major components on Replica in \tref{tab:abl_score_merge}. Disabling point-level confidence updating degrades overall mapping performance and increases Duplication (measured as the average number of mapped instances matched to each ground-truth object) from 1.1690 to 1.2568, as integrated outlier points and false instances can no longer be suppressed by contradictory observations. Disabling fragment merging further increases Duplication to 1.3426, as fragments arising from disconnected observations can no longer be merged.

\begin{figure}[!t]
    \centering
    \includegraphics[width=0.9\linewidth]{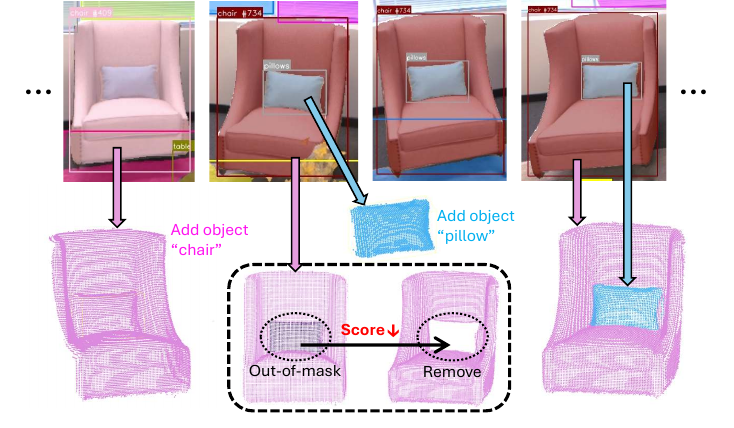}
    \caption{
    \textbf{Example of handling under-segmentation.} A pillow is initially fused into the chair instance due to under-segmentation. Once the pillow is correctly segmented in later frames, our method adds it as a separate instance and gradually removes the corresponding points from the chair instance through confidence updates.
    }
    \label{fig:sofa}
\end{figure}

\begin{figure}[!t]
    \centering
    \includegraphics[width=0.8\linewidth]{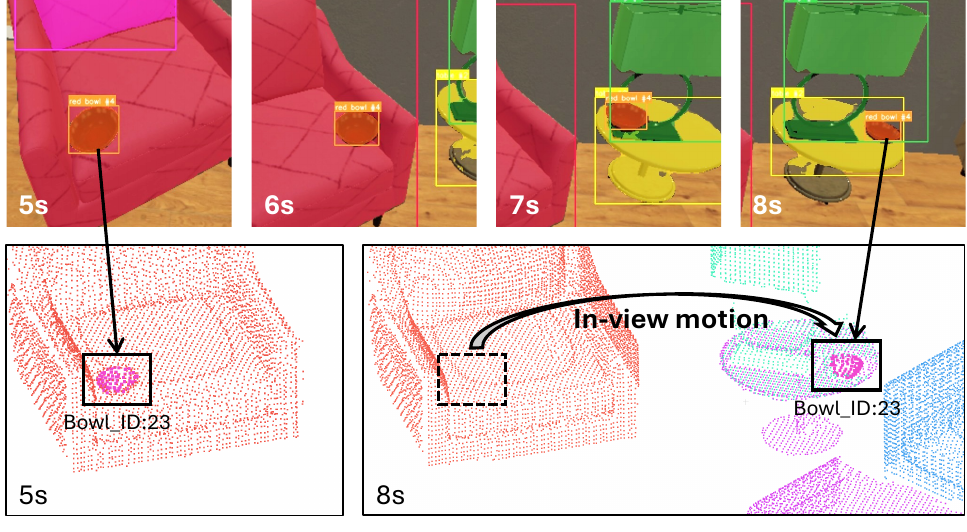}
    \caption{
    \textbf{Example of in-view object tracking.} Our method tracks the moving bowl in images and updates its location in the semantic map while preserving the same instance ID.
    }
    \label{fig:bowl}
\end{figure}

\subsection{Robustness to Segmentation Errors}

To evaluate the robustness of our point-level probabilistic updates to segmentation errors, we conduct a perturbation test following \cite{heller2018imperfect}. Specifically, we add Gaussian noise $\mathcal{N}(0,\sigma=3\,\text{px})$ to the predicted mask boundaries and evaluate both our method and DualMap on the perturbed masks. We report the performance drop due to such perturbations in \tref{tab:drop}.
Our method shows smaller degradation under mask errors. This is because our probabilistic confidence update mechanism suppresses the points introduced by segmentation errors based on temporal and geometric consistency. The improvement is more pronounced in mIoU and mAcc than in F-mIoU, since F-mIoU gives higher weight to large objects, where boundary outliers occupy only a small fraction.

\subsection{Qualitative Analysis of Common Mapping Errors}
We analyze four common error sources in open-vocabulary semantic mapping and show how our system mitigates them.

\subsubsection{Inaccurate Masks}
As shown in \fref{fig:chair}, segmentation masks for thin and hollow structures may incorrectly fill empty regions, introducing outlier points into the 3D object map. DualMap retains these mask-induced artifacts once they are fused, while our method suppresses inconsistent points across views and reconstructs cleaner geometry.

\subsubsection{Incorrect Detections}
As shown in \fref{fig:plastic}, a plastic bag is mistakenly detected as a pillow in some frames, creating a false instance in the map. Our method removes this false instance through confidence updates while preserving the true pillow, showing robustness to incorrect detections.

\subsubsection{Object Fragmentation}
As shown in \fref{fig:table}, a table is initially reconstructed as two separate instances because its two sides are observed at different times with limited overlap. Once more complete observations become available, our method merges these fragments into a single instance.

\subsubsection{Under-segmentation}
As shown in \fref{fig:sofa}, a pillow is initially fused into the chair instance due to under-segmentation. When the pillow is correctly segmented in later frames, our method adds it as a separate instance and removes the corresponding erroneous points from the chair.

\begin{table}[!t]\centering\small
\caption{Comparison of out-of-view object relocation.}\label{tab:reid_comp}
\resizebox{\linewidth}{!}{
\begin{tabular}{lccccc}\toprule
\textbf{Method} &\textbf{Moved Objs.} &\textbf{Found Old} &\textbf{Removed Old} &\textbf{Found New} &\textbf{Succ. ReID} \\\midrule
DualMap &\multirow{2}{*}{26} &12 (46.2\%) &10 (38.5\%) &15 (57.7\%) &-- \\
\textbf{Ours} & &\textbf{26 (100\%)} &\textbf{26 (100\%)} &\textbf{24 (92.3\%)} &\textbf{20 (76.9\%)} \\
\bottomrule
\end{tabular}
}
\end{table}

\begin{figure}[!t]
\centering
\includegraphics[width=\linewidth]{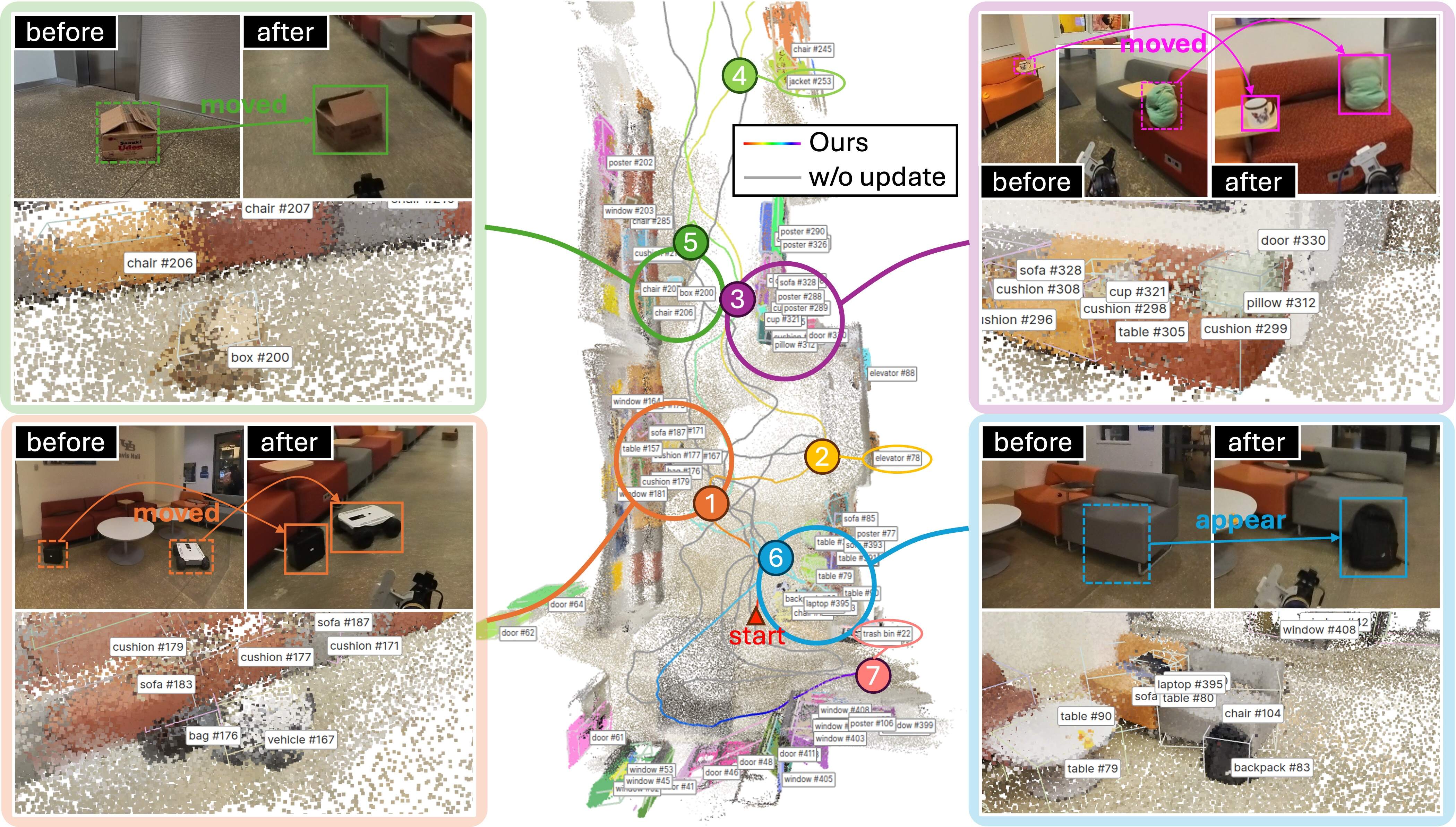}
\caption{\textbf{Real-world VLN evaluation under environment changes.} After objects are moved, the agent executes 7 object navigation tasks in sequence. Our full system travels 76.41 m, whereas the ablation without map refinement travels 103.22 m because it revisits stale object locations.}
\label{fig}
\end{figure}

\subsection{Dynamic Object Handling}
We evaluate dynamic object handling in five HSSD scenes \cite{khanna2023hssd} with movable objects from the YCB dataset \cite{calli2017yale}.

\subsubsection{Out-of-View Relocation and Re-Identification}

We compare our method with DualMap under identical observation sequences before and after object relocation. For DualMap, we evaluate its global abstract map.
As shown in \tref{tab:reid_comp}, DualMap only retains 46.2\% and 57.7\% of the objects before and after object relocation (Found Old/New), because its handling of dynamic objects relies on establishing relations to relatively stable anchors, which do not always exist. In contrast, our relation-free representation retains 100\% of the original objects and 92.3\% after relocation; the missing two objects become too small to be detected at their new locations. For stale-object removal, DualMap only removes 38.5\% of the old instances (Removed Old) because its updates require sufficiently complete observation of the associated anchor, whereas our method performs point-level visibility reasoning directly on the object itself and removes all 26 stale instances. Our re-identification module further recovers the original identity for 20 of the 26 relocated objects (76.9\%); four objects with ambiguous matches are conservatively rejected.

\subsubsection{In-View Tracking}
In this experiment, objects move while remaining visible. \fref{fig:bowl} shows an example where a bowl is moved from a chair to a table. The semantic map updates the same instance over time, preserving its ID while reflecting its new location.
Quantitatively, we evaluate on sequences of 32 different objects. Our method successfully tracks and updates 27 objects, achieving a success rate of 84.4\%. The remaining 5 cases fail because the objects become too small to detect when moved farther away.

\begin{figure}[!t]
    \centering
    \includegraphics[width=\linewidth]{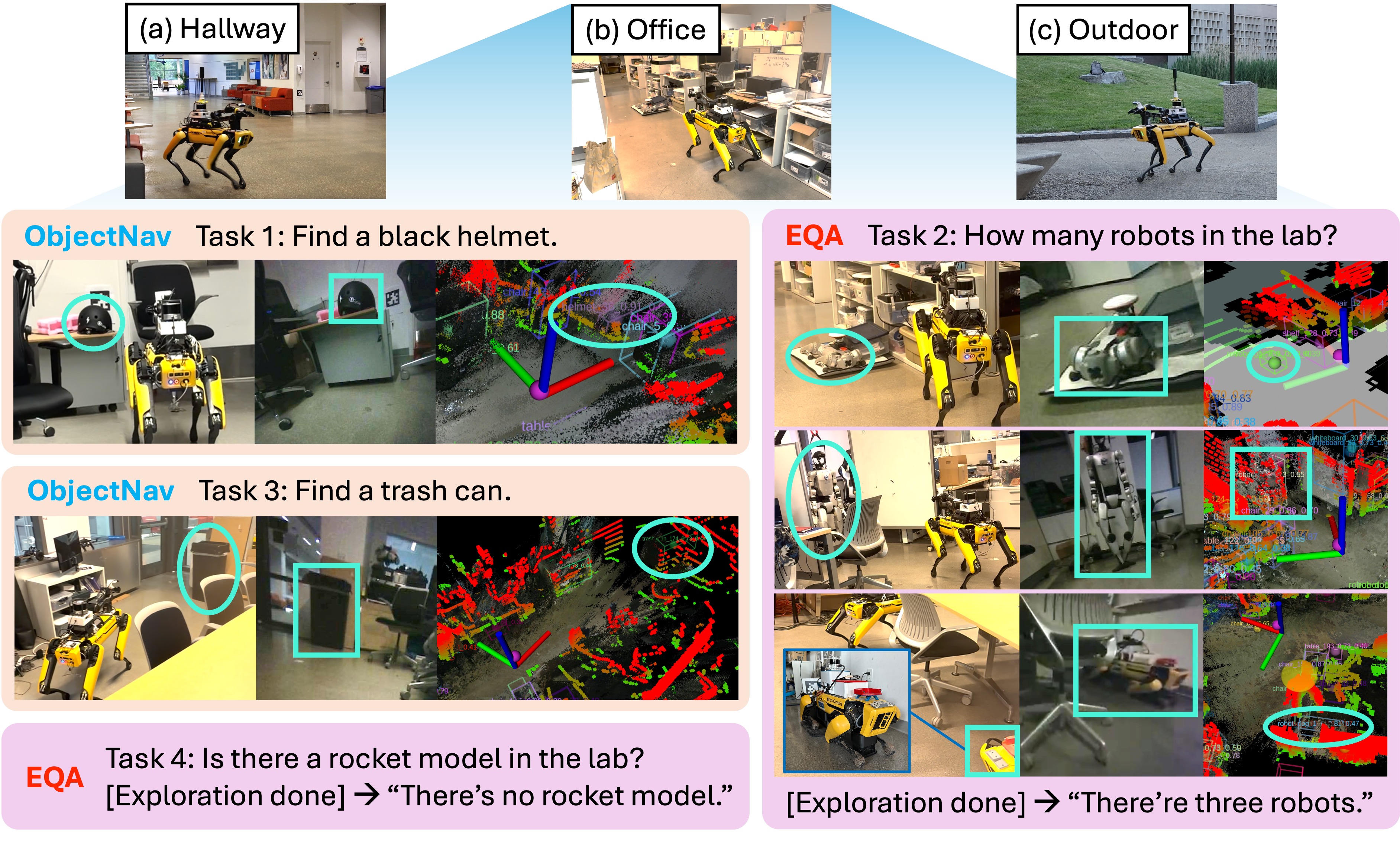}
    \caption{
    Examples of object navigation and EQA tasks executed on the Spot robot. Our semantic map serves as an online memory that helps the LLM localize target objects, plan embodied skills, and answer questions.
    }
    \label{fig:real}
\end{figure}

\begin{table}[!t]
\centering
\small
\caption{Performance on real-world object navigation and embodied question answering tasks.}
\label{tab:real_tasks}
\resizebox{\linewidth}{!}{
\begin{tabular}{l|ccc|ccc}
\toprule
& \multicolumn{3}{c|}{\textbf{Object Navigation}} 
& \multicolumn{3}{c}{\textbf{EQA}} \\
\textbf{Scenes} 
& \textbf{Trials} & \textbf{SR \% ($\uparrow$)} & \textbf{SPL ($\uparrow$)}
& \textbf{Trials} & \textbf{SR \% ($\uparrow$)} & \textbf{SPL ($\uparrow$)} \\
\midrule
Hallway & 11 & 72.7 & 0.603 & 9  & 66.7 & 0.532 \\
Office  & 12 & 91.7 & 0.790 & 12 & 83.3 & 0.714 \\
Outdoor & 11 & 81.8 & 0.685 & 9  & 77.8 & 0.641 \\
\midrule
\textbf{All} & \textbf{34} & \textbf{82.4} & \textbf{0.696}
             & \textbf{30} & \textbf{76.7} & \textbf{0.638} \\
\bottomrule
\end{tabular}
}
\end{table}

\subsection{Real-World Evaluation on VLN Tasks} \label{sec:real_vln}

We evaluate the proposed semantic map as online spatial memory for real-world VLN tasks.
Experiments are conducted on a Boston Dynamics Spot robot with RGB-D cameras and LiDAR-based localization.

\subsubsection{Benefit of Map Refinement}

We first evaluate whether map refinement benefits downstream tasks when the environment changes. An initial semantic map is built from a teleoperated sequence, after which the objects are rearranged. The robot then autonomously executes 7 sequential object navigation tasks using the original map as initialization. We compare the full system with an ablation that disables map refinement.
Both systems complete all 7 tasks, but the full system travels only 76.41 m compared with 103.22 m without map refinement. Continuous map updates remove stale objects during execution, preventing subsequent tasks from repeatedly selecting and revisiting obsolete object locations.

\subsubsection{Broader VLN Evaluation}

We further evaluate the complete system in three unseen environments, \textit{Hallway}, \textit{Office}, and \textit{Outdoor}. We consider both object navigation and embodied question answering (EQA). For object navigation, the LLM queries the map for matching object instances and their 3D locations, while continuing exploration if the target has not yet been observed. For EQA, the LLM retrieves relevant object instances, semantic labels, locations, and spatial context, and explores further when the current map is insufficient to answer the question. Representative examples are shown in \fref{fig:real}. As reported in \tref{tab:real_tasks}, the robot achieves an average success rate of 82.4\% for object navigation and 76.7\% for EQA, demonstrating that the proposed map can serve as effective online spatial memory for both goal-directed navigation and semantic-spatial reasoning in previously unseen environments.

\section{Conclusion}
\label{sec:conclusion}

We presented a dynamic open-vocabulary semantic mapping framework that turns object-centric maps from append-only accumulators into point-level probabilistic representations. By assigning each object point a log-odds confidence score and updating it with multi-view observation evidence, our method reinforces stable points while suppressing perception artifacts and removing old points caused by object motion. It also preserves moved-object identities through tracking and re-identification. 
Experiments on Replica, ScanNet, and dynamic-object scenarios show improved semantic mapping accuracy, robustness to perception errors, and effective handling of moved or removed objects. Real-world navigation and embodied question-answering experiments further validate the proposed map as an online spatial memory for language-guided robotic agents.


\balance
\bibliographystyle{IEEEtran}
\bibliography{references}

\end{document}